\documentclass[10pt,twocolumn,letterpaper]{article}

\usepackage{iccv}
\usepackage{times}
\usepackage{epsfig}
\usepackage{graphicx}
\usepackage{amsmath}
\usepackage{amssymb}

\usepackage[breaklinks=true,bookmarks=false]{hyperref}

\iccvfinalcopy 

\def\iccvPaperID{1395} 

\ificcvfinal\fi

\begin{document}

\title{Differentiable Learning-to-Group Channels via \\ Groupable Convolutional Neural Networks}
\author{\normalsize
Zhaoyang Zhang\textsuperscript{\scriptsize{1}} \quad Jingyu Li\textsuperscript{\scriptsize{2}}\quad Wenqi Shao\textsuperscript{\scriptsize{1}}\quad Zhanglin Peng\textsuperscript{\scriptsize{2}} \quad Ruimao Zhang\textsuperscript{\scriptsize{2}}\quad Xiaogang Wang\textsuperscript{\scriptsize{1}}\quad Ping Luo\textsuperscript{\scriptsize{3}}  \\
{\small \textsuperscript{\scriptsize{1}}CUHK-SenseTime Joint Laboratory, The Chinese University of Hong Kong
 \qquad \textsuperscript{\scriptsize{2}}SenseTime Research \qquad \textsuperscript{\scriptsize{3}} The Univesity of Hong Kong}\\
{\tt\small \{zhaoyangzhang@link, weqish@link, xgwang@ee\}.cuhk.edu.hk}  \\
{\tt\small \{lijingyu, pengzhanglin, zhangruimao\}@sensetime.com} \quad
{\tt\small pluo@cs.hku.hk}
}

\maketitle
\ificcvfinal\thispagestyle{empty}\fi

\begin{abstract}
Group convolution, which divides the channels of ConvNets into groups, has achieved impressive improvement over the regular convolution operation. However, existing models, \eg ResNeXt, still suffers from the sub-optimal performance due to manually defining the number of groups as a constant over all of the layers. Toward addressing this issue, we present Groupable ConvNet (GroupNet) built by using a novel dynamic grouping  convolution (DGConv) operation, which is able to learn the number of groups in an end-to-end manner. The proposed approach has several appealing benefits. (1) DGConv provides a unified convolution representation and covers many existing convolution operations such as regular dense convolution, group convolution, and depthwise convolution. (2) DGConv is a differentiable and flexible operation which learns to perform various convolutions from training data. (3) GroupNet trained with DGConv learns different number of groups for different convolution layers. Extensive experiments demonstrate that GroupNet outperforms its counterparts such as ResNet and ResNeXt in terms of accuracy and computational complexity. We also present introspection and reproducibility study, for the first time, showing the learning dynamics of training group numbers.
\end{abstract}

 \section{Introduction}
 Convolutional Neural Networks (ConvNets) have achieved remarkable successes in computer vision.
    For example, ResNet ~\cite{he2016deep} was a pioneer work on building very deep networks with shortcut connections.
    This strategy exposes depth of network as an essential dimension of ConvNets to achieve good performance.
    Other than tailoring network architectures on depth and width~\cite{szegedy2015going,ioffe2015batch,szegedy2017inception,szegedy2016rethinking}, ResNeXt~\cite{xie2017aggregated} proposed a new dimension ``cardinality'', utilizing group convolution to design effective and efficient ConvNets.
    The main hallmark of group convolution is proven to be compact and parameter-saving, which means that ResNeXt improves accuracy and reduces network parameters, outperforming its counterpart ResNet.

\begin{figure}[t]
    \begin{center}
    \includegraphics[width=1.0\linewidth]{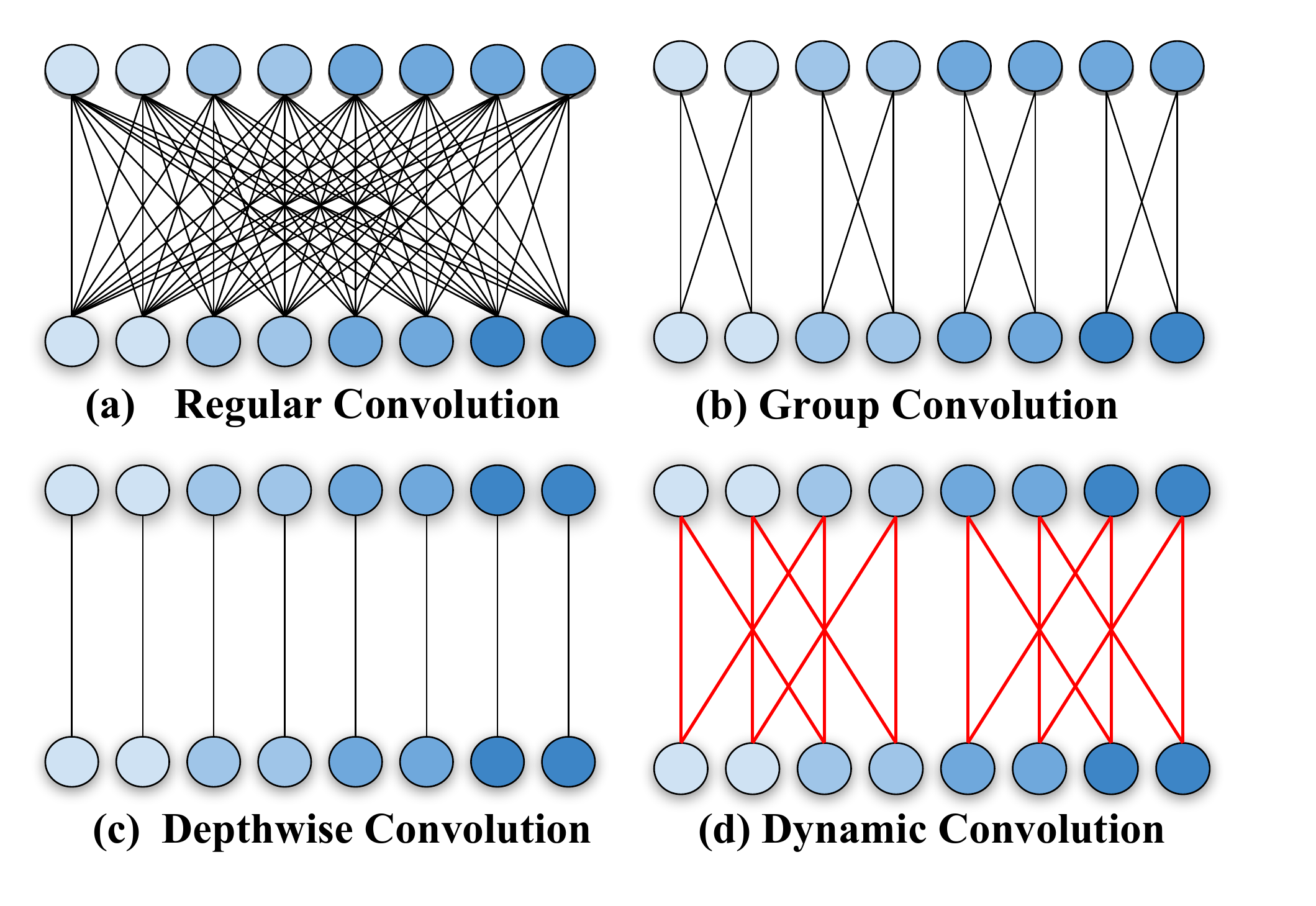}
    \caption{Illustration of different convolution strategies, where the blue circles represent input and output channels, and the lines are the connections between them. (a) Regular convolution. Every input channel is connected to every output channel. (b) Group convolution with cardinality $4$ and width $2$. (c) Depthwise convolution. Each input channel is connected to only one output channel, so this convolution can be understood as linear transformation for each channel. (d) Our proposed dynamic grouping convolution (DGConv). \emph{The grouping  strategy of DGConv is learned end-to-end together with the network parameters , so the group number and connection location are changing dynamically.} This example is one candidate strategy with 2 groups and non-adjacent channel connection. During the test stage, the DGConv can be simply implemented by group convolution with the group number learned from training, which reduces the computations and parameters.
}
    \label{fig:1}
    \end{center}
\end{figure}

    Although group convolution is easy to implement, applying group convolution in previous networks such as ResNeXt still has drawbacks.

    First, when designing network architectures by using group convolutions, the number of groups for each hidden layer has been treated as a hyper-parameter typically. The group number is often defined by human experts and kept the same for all hidden layers of a ConvNet.
%
    Second, previous work employed homogeneous group convolutions, leading to sub-optimal solution. For instance, one of the most practical setting of ResNeXt is ``32x4d'' that applies group convolution with 32 groups, which is found by trial and error. However, convolution layers in different depths of a ConvNet typically learn different visual features which represent different abstractions and semantic meanings. Thus, uniformly reducing model parameters via group convolutions may suffer from decreasing performance.
%

    To address the above issues, this work introduces an autonomous formulation of group convolution, naming \emph{Dynamic Grouping Convolution (DGConv)}, which generally extends many convolution operations with the following appealing properties. (1) \textbf{Dynamic grouping .} The core of DGConv is to train the convolution kernels and the grouping  strategy simultaneously. As shown in Fig.~\ref{fig:1}, DGConv is able to learn grouping  strategy (\ie group number and connections between channels in a group) during training. In this way, each DGConv layer can have individual grouping  strategy.  Moreover, by imposing a regularization term on computational complexity, we can control the overall model size and computational overhead.
    (2) \textbf{Differentiability.} The learning of DGConv is fully differentiable and can be trained in an end-to-end manner by using stochastic gradient descent (SGD). Thus, DGConv is compatible with existing ConvNets.
    %
    (3) \textbf{Parameter-saving.} The extra parameters to learn the grouping  strategy in DGConv is just scaled in $\log_2(C)$, where $C$ is the number of channels of a convolution layer.
    This extra number of parameters is far less than the parameters of the convolution kernels, which are proportional to the scale\footnote{The kernel parameters are $C^{\mathrm{in}}\times C^{\mathrm{out}}$, which indicate the input channel size and the output channel size.} of $C^2$. 
    
\begin{figure}[t]
    \begin{center}
   \includegraphics[width=1.0\linewidth]{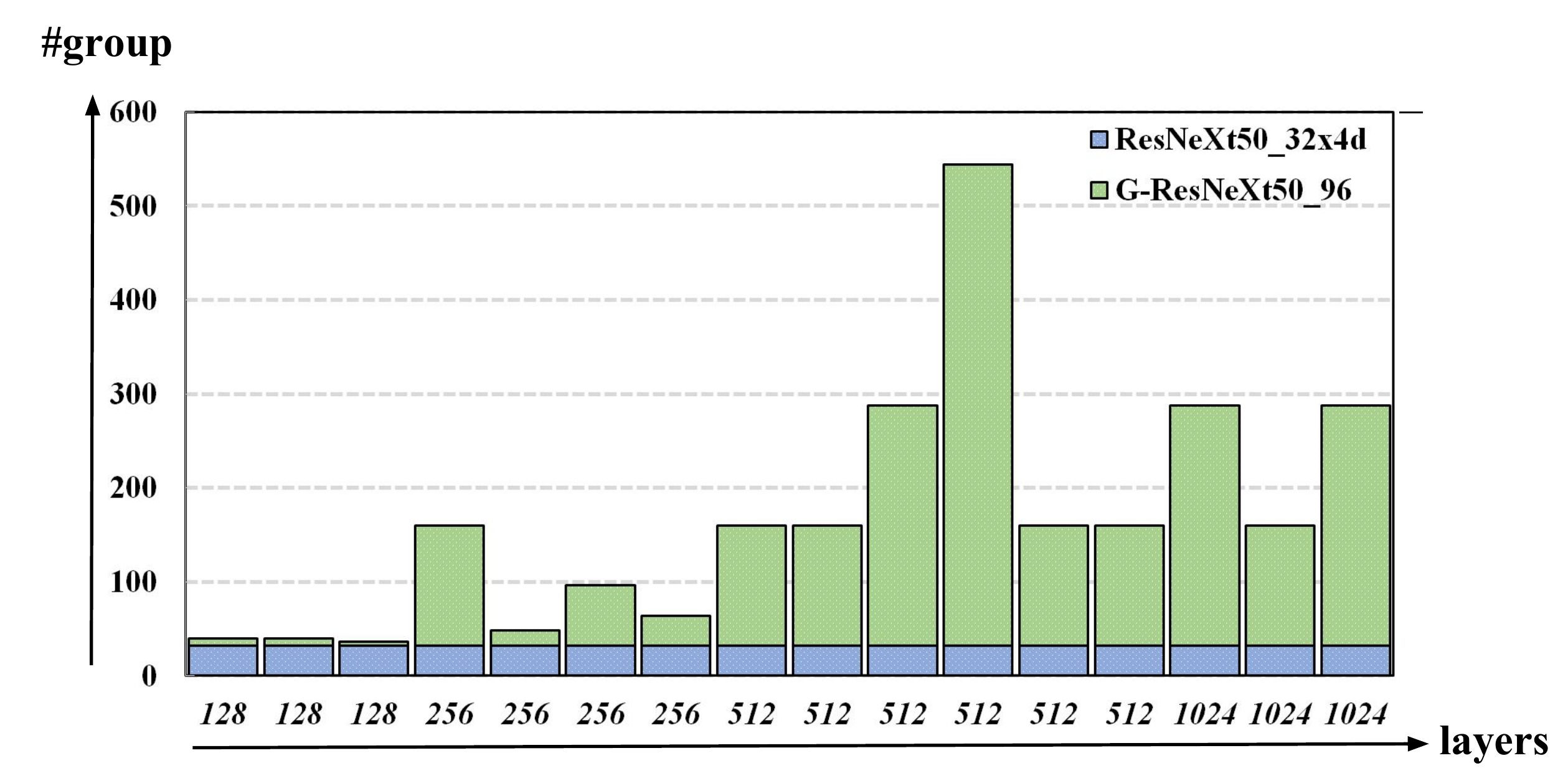}
    \caption{Comparison of group numbers in ResNeXt and GroupNet. We employ \textit{ResNeXt50$\_$32$\times$4d} as an example here, which has 32 groups with width 4. And \textit{G-ResNeXt50$\_$96} denotes the ResNeXt50 trained with DGConv, where $96$ represents the constraint setting (will be discussed later).
     The y-axis indicates number of groups, and the x-axis is the number of channels in different convolution layers.
     }
    \label{fig:2}
\end{center}
\end{figure}

    Furthermore, the extra parameters could be discarded after training. In the testing stage, only the parameters of the convolution kernels will be stored and loaded. Fig.~\ref{fig:2} shows an example of the group numbers learned by DGConv, which is able to achieve comparable performance with respect to its counterpart, but significantly reducing parameters and computations.

    This work makes three key \textbf{contributions}. (1) We propose a novel convolution operation, \emph{Dynamic Grouping Convolution (DGConv)}, which is able to \emph{differentiably} learn the number of groups for group convolution, unlike existing work that treated the group number as a hyper-parameter. To our knowledge, this is the first time to learn group number in a differentiable and data-driven way.
    (2) DGConv can be used to replace previous convolutions and build state-of-the-art deep networks such as the proposed Groupable ResNeXt in section~\ref{sec:3.3}, where the group number for each convolution layer is automatically determined during end-to-end training.
%
    (3) Extensive experiments demonstrate that Groupable ResNeXt is able to outperform both ResNet and ResNeXt, by using comparable or even smaller number of parameters.
    For example, it surpasses ResNeXt101 by 0.8\% top-1 accuracy in ImageNet with slightly less parameters and computations.
%
    Moreover, we study the learning dynamics of group numbers, showing interesting findings.

\section{Related Work}

\textbf{Group Convolution.} Group convolution (GConv) is a special case of sparsely connected convolution.
In regular convolution, we produce $C^{\mathrm{out}}$ output channels by applying convolution filters over all $C^{\mathrm{in}}$ input channels, resulting in a computational cost of $C^{\mathrm{in}} \times C^{\mathrm{out}}$.
In contrast, GConv reduces this cost by dividing the $C^{\mathrm{in}}$ input channels into $G$ non-overlapping groups. After applying filters over each group, GConv generates $C^{\mathrm{out}}$ output channels by concatenating the outputs of each group. GConv has a complexity of $\frac{C^{\mathrm{in}} \times C^{\mathrm{out}}}{G}$.

GConv is firstly discussed in AlexNet \cite{krizhevsky2012imagenet} as a model distributing approach to handle memory limitation.
ResNeXt \cite{xie2017aggregated} presented an additional dimension for network architecture \ie ``cardinality'' by using GConv,
leading to a series of further researches on applying group convolution in portable neural architecture design~\cite{zhang2017interleaved, ma2018shufflenet, zhang2018shufflenet, huang2018condensenet}.
To the extreme, group convolution partitions each channel into a single group, which is known as depthwise convolution. It has been widely used in efficient neural architecture design \cite{howard2017mobilenets, ma2018shufflenet, zhang2018shufflenet, sandler2018mobilenetv2}.

Moreover, CondenseNet~\cite{huang2018condensenet} and FLGC~\cite{Wang_2019_CVPR} learned the connections of group convolution, but the number of groups is still a predefined hyper-parameter.
%
%
CondenseNet and FLGC treated connection learning as a pruning problem, where unimportant filters are abolished.
In contrast, DGConv learns both the group number and the channel connections of each group.
%


\textbf{Neural Architecture Search.} Recently, there has been growing interests in automating the design process of neural architectures, usually referred as Neural Architecture Search (NAS) and AutoML.
For example, NASNet~\cite{zoph2018learning, zoph2016neural} and MetaQNN~\cite{baker2016designing} lead the trend of architecture search by using reinforcement learning (RL).
In NASNet, the network architecture is decomposed into repeatable and transferable blocks, such that the control parameters of the architectures can be limited in a finite searching space.
The sequence of these architecture parameters was generated by a controller RNN, which is trained by maximizing rewards (\eg val accuracy).
These methods were extended in many ways such as progressive searching \cite{liu2018progressive}, parameter sharing \cite{pham2018efficient}, network transformation \cite{cai2017reinforcement}, resource-constrained searching \cite{tan2018mnasnet}, and differentiable searching like DARTS \cite{liu2018darts} and SNAS~\cite{xie2018snas}.
Evolutionary algorithm is an alternative to RL. The architectures are searched by mutating the best architectures found so far~\cite{real2018regularized, real2017large, xie2017genetic, miikkulainen2019evolving,liu2017hierarchical}.
However, all the above methods either treated the group number as a hyper-parameter, or searched its value by using sampling methods such as RL. In contrast, DGConv is the first model that can optimize the group number in a data-driven way and a differentiable end-to-end manner together with the network parameters.

\begin{figure*}[t]
\begin{center}
   \includegraphics[width=0.85\linewidth]{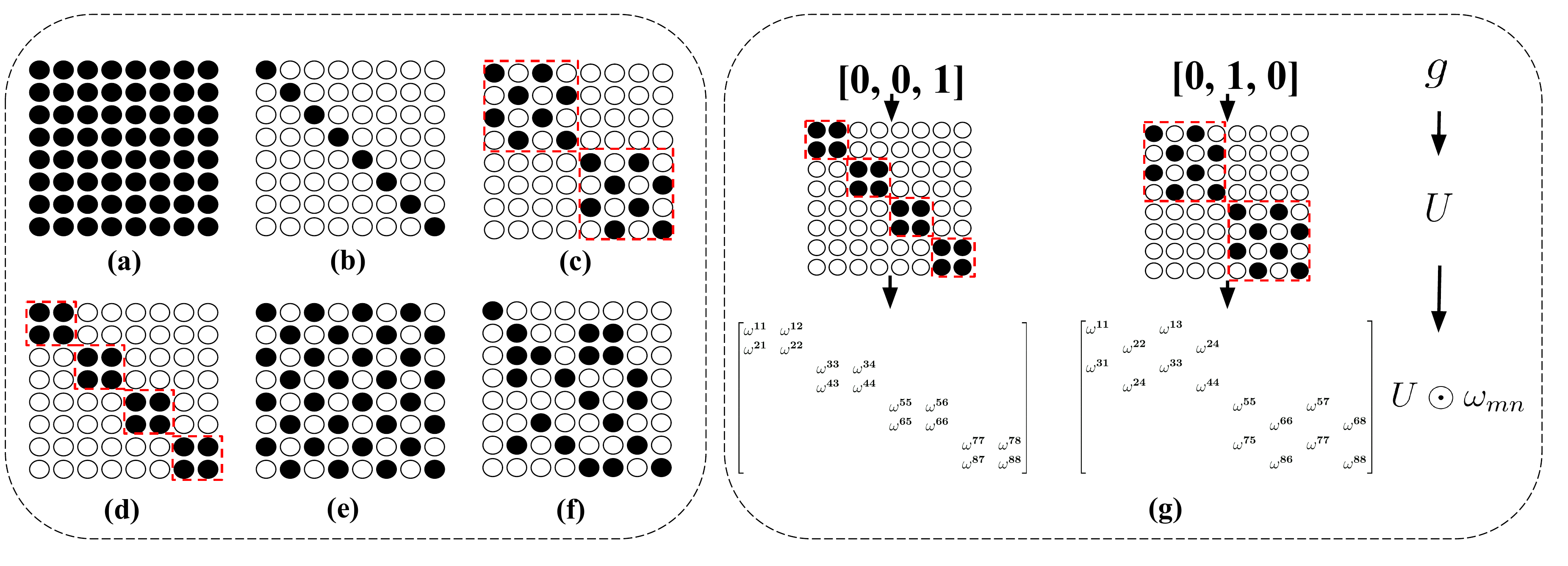}
    \caption{\textbf{Illustration of structures with relationship matrix $\mathbf{U}$.} The hollow circle and solid black circle indicate `$0$' and `$1$' respectively. A matrix of ones(a), identity matrix(b) and block diagonal matrix(d) imply regular convolution, depthwise convolution and group convolution (GConv) respectively. (c) and (e) show Dynamic Grouping Convolution (DGConv) under two non-adjacent group strategies respectively, one with a group number of 4 and the other with 2. (f) is a random group strategy, while it cannot been achieved under our constraint. (g) illustrates the construct process of DGConv when $g=[0, 0, 1]$ and $g=[0, 1, 0]$. The binary relationship matrix $U$ disables weights of $\omega$ via elementwise product operation.}
    \label{fig:3}
\end{center}
\end{figure*}

\section{Our Approach}

\subsection{Dynamic Grouping Convolution (DGConv)}

We first present conventional convolution and group convolution, and then introduce DGConv.

\textbf{Regular Convolution.} Let a feature map of a ConvNet be~$F\in \mathbb{R}^{N\times C^{\mathrm{in}}\times H\times W}$, where $N,C,H,W$ represent number of samples in a minibatch, number of channels, height and width of a channel respectively.
If a regular convolution is applied on $F$ with kernel size $k \times k$ and stride $1$ with padding, the output feature map is denoted as $O\in \mathbb{R}^{N\times C^{\mathrm{out}}\times {H}\times {W}}$, where every output unit~$o_{ij}\in\mathbb{R}^{N\times C^{\mathrm{out}}}$ is
\begin{equation}
    \label{eq1}
    o_{ij} = \sum_{m=0}^{k-1} \sum_{n=0}^{k-1}  f_{(i+m)(j+n)}\omega_{mn},
\end{equation}
where~$i\in\{1,...,H\}$, $j\in\{1,...,W\}$, and $f_{(i+m)(j+n)} \in \mathbb{R} ^ {N \times C^{\mathrm{in}}}$ represents the hidden units of the input feature map $F$. And $\omega_{mn} \in \mathbb{R} ^ {C^{\mathrm{in}} \times C^{\mathrm{out}}}$ represents the convolution weights (kernels).

\textbf{Group Convolution.} Group convolution (GConv) can be defined as a regular convolution with sparse kernels. GConv is often implemented as concatenation of separated convolution over grouped channels,
\begin{equation}
    \label{eq2:group}
           \begin{matrix}
           o_{ij}=o_{ij}^{1} \cup \dots \cup o_{ij}^{\gamma} \cup \dots \cup o_{ij}^{G}~~ \mathrm{and}~~\\
           o_{ij}^{\gamma} = \sum_{m=0}^{k-1} \sum_{n=0}^{k-1} f_{(i+m)(j+n)}^{\gamma} \omega_{mn}^{\gamma},
           \end{matrix}
\end{equation}
where $G$ is the group number, $\gamma \in [1, G]$, and $\cup$ means the concatenation operation.
In context of GConv, we have $\omega^\gamma_{mn} \in \mathbb{R} ^ {\frac{C^{\mathrm{in}}}{G}\times \frac{C^{\mathrm{out}}}{G}}$ and $f^\gamma_{(i+m)(j+n)} \in \mathbb{R} ^ {N \times \frac{C^{\mathrm{in}}}{G}}$.
To the extreme, when every channel is a group \ie $G=C^{\mathrm{in}}=C^{\mathrm{out}}$, Eqn.\eqref{eq2:group} expresses the depthwise convolution ~\cite{howard2017mobilenets, sandler2018mobilenetv2, ma2018shufflenet, zhang2018shufflenet}.
Both GConv and depthwise convolution reduce computational resources and can be efficiently implemented in existing deep learning libraries.
However, intrinsic hyper-parameter G is manually designed, making performance away from idealism.

\textbf{Dynamic Grouping Convolution.} Dynamic grouping  convolution (DGConv) extends group convolution, enabling to learn grouping  strategies, that is, group number and channel connections of each group.
The strategies can be modeled by a binary relationship matrix~$U\in \{0, 1\}^{C^{\mathrm{in}} \times C^{\mathrm{out}}}$.
DGConv can be defined as
\begin{equation}
    \label{eq2}
    o_{ij} = \sum_{m=0}^{k-1} \sum_{n=0}^{k-1}f_{(i+m)(j+n)}(U\odot \omega_{mn}),
\end{equation}
where $\odot$ denotes elementwise product.
It is note-worthy that Eqn.\eqref{eq2} has rich representation capacity.
Many convolution operations can be treated as special cases of DGConv.
To build some intuition on flexibility of DGConv, several illustrative examples are presented in the following:

(1) Let $U=\mathbf{1}$, where $\mathbf{1}$ is a matrix of ones. Since we have $\mathbf{1}\odot \omega_{mn}=\omega_{mn}$, DGConv represents a regular convolution, as shown in Fig.~\ref{fig:3} (a).
(2) Let~$U = I$, where $I$ is an identity matrix. Then $I\odot \omega_{mn}$ becomes a matrix with diagonal elements while the off-diagonal elements are zeros as depicted in Fig.~\ref{fig:3} (b), implying that every channel is independent. Thus, DGConv becomes a depthwise convolution~\cite{howard2017mobilenets}.
(3) If $U$ is a binary block-diagonal matrix as shown in Fig.~\ref{fig:3} (d), then $U\odot \omega_{mn}$ divides channels into groups. Since all diagonal blocks of $U$ are constant matrix of ones, DGConv expresses a conventional group convolution (GConv), which groups adjacent channels as a group.
(4) If $U$ is an arbitrary binary matrix such as Fig.~\ref{fig:3} (f), this leads to unstructured convolution.

Therefore, by appropriately constructing binary relationship matrix $U$,the  proposed DGConv is expected to represent a large variety of convolution operations.

\textbf{Discussions.} We have defined DGConv as above. Although it has huge potential to boost learning capacity of CNN due to its flexibility in convolution representation, some foreseeable difficulties are also introduced.

First, since Stochastic Gradient Descent (SGD) can only optimize continuous variables, training a binary matrix  by directly using SGD can be challenging.
Second, the matrix $U\in\{0,1\}^{C^{\mathrm{in}} \times C^{\mathrm{out}}}$ introduces a large amount of extra parameters into the convolution operation, making the deep networks difficult to train.
Third, updating the entire matrix $U$ without any constraint in the training stage could learn a unstructured relationship matrix $U$ as illustrated in Fig.~\ref{fig:3} (f). In this case, DGConv is not a valid GConv, making learned convolution operation inexplicable.

Therefore, for DGConv, special construction of $U$ is required to maintain the group structures and reduce the extra number of parameters.

     %


\subsection{Construction of the Relationship Matrix}


Instead of directly learning the entire matrix $U$, we decompose it into a set of $K$ small matrixes,
\begin{equation*}
        \{ U_k  | U_k \in \{0, 1\}^{C_k^{\mathrm{in}} \times C_k^{\mathrm{out}}}, \forall C_k^{\mathrm{in}} < C^{\mathrm{in}}, \forall C_k^{\mathrm{out}} < C^{\mathrm{out}}\}.
\end{equation*}
We see that each small matrix $U_k$ is of shape $C_k^{\mathrm{in}} \times C_k^{\mathrm{out}}$, where $C_k^{\mathrm{in}} < C^{\mathrm{in}}$ and $C_k^{\mathrm{out}} < C^{\mathrm{out}}$.
%
%
Then we define $U$ as
\begin{equation}
        \label{eq3}
        U = U_1 \otimes U_2 \otimes \dots  \otimes U_K,
\end{equation}
where $\otimes$ denotes a Kronecker product. Therefore, we have $\prod_{k=1}^{K}C_k^{\mathrm{in}} = C^{\mathrm{in}}$ and $\prod_{k=1}^{K}C_k^{\mathrm{out}} = C^{\mathrm{out}}$, implying that the $C^{\mathrm{in}}$-by-$C^{\mathrm{out}}$ large matrix $U$ is decomposed into a set of small submatrixes by using a sequence of Kronecker products~\cite{batselier2017constructive}.

\textbf{Construction of Submatrix.} Here we introduce how to construct each submatrix $U_k$. As an illustrative example, we suppose $C^{\mathrm{in}}=C^{\mathrm{out}}$, which is a common setting in ResNet and ResNeXt.
To pursue a most parameter-saving convolution operation, we further represent $U_k$ by a single binary variable as follow:
%
\begin{equation}\label{eq6}
            \left\{\begin{array}{rl}  U_k &= {g}_k\mathbf{1} + (1 -{g}_k) I , ~~\forall {g}_k \in {g},\\
            {g} &= \mathrm{sign}(\tilde{g}),
                        \end{array}\right.
\end{equation}
where $\mathbf{1}$ denotes a 2-by-2 constant matrix of ones, $I$ denotes a 2-by-2 identity matrix and $g_k$ indicates the $k$-th component. $\tilde{g} \in \mathbb{R} ^ {K}$ is a learnable gate vector taking continues value, and ${g} \in \{0, 1\}^{K}$ is a binary gate vector derived from $\tilde{g}$.
The $\mathrm{sign}(\cdot)$ represents a sign function,
\begin{equation}
\label{eq4}
       \mathrm{sign}(x) = \left\{\begin{matrix} 0,~~~x < 0.
                                    \\ 1, ~~~x \geqslant 0.
                        \end{matrix}\right.
    \end{equation}

By combing Eqn.\eqref{eq6}, Eqn.\eqref{eq3} could be written as
\begin{equation}
        \label{eq7}
         U =({g}_1 \mathbf{1} + (1 -{g}_1) I) \otimes \dots  \otimes ({g}_K \mathbf{1} + (1 -{g}_K) I).
\end{equation}
Constructing relationship matrix $U$ by Eqn.\eqref{eq7} not only remarkably reduces the amount of parameters but also makes $U$ have group structure. First, note that the parameters to be optimized are $\tilde{g}$, the above construction method therefore reduces the number of parameters of $U$ from $C^{\mathrm{in}}\cdot C^{\mathrm{out}}$ to $\log_2C^{\mathrm{in}}$.
For example, if there is $1,024$ channels of a convolution layer, we can learn the block diagonal matrix $U$ in Eqn.\eqref{eq7} by using merely 10 parameters, remarkably reducing the number of training parameters, which previously is more than $10^6$.
Second, we see that $U$ constructed by Eqn.\eqref{eq7} is a symmetric matrix with diagonal element of ones. Moreover, each row or column of  $U$ has the same elements. Hence, $U$ has a group structure.
For example, when $K=3$ and $g_1=1,g_2=1,g_3=0$, Eqn.\eqref{eq7} becomes $\mathbf{1} \otimes \mathbf{1} \otimes I$, which is a 8-by-8 matrix of 2 groups as shown in Fig.~\ref{fig:3} (e); when $g_1=0,g_2=1,g_3=0$, Eqn.\eqref{eq7} becomes $I \otimes \mathbf{1} \otimes I$, which is a 8-by-8 matrix of 4 groups as shown in Fig.~\ref{fig:3} (c).
They show that our proposed DGConv can group non-adjacent channels. Fig.~\ref{fig:3} (g) shows the dynamical process of actual of DGConv when $g=[0,0,1]$ and $g=[0,1,0]$.
It can be observed that the position of `$1$' in $g$ can control the group structure of $U$ and $U\odot\omega_{mn}$.
Note that we use only 3 continuous parameters $\tilde{g}_1,\tilde{g}_2,\tilde{g}_3$ to produce $g_1,g_2,g_3$, enabling to learn the large 8-by-8 matrix that originally needs 64 parameters to train.
A more general case when $C^{\mathrm{in}}\neq C^{\mathrm{out}}$ is discussed in Appendix A.

\textbf{Training Algorithm of DGConv.}
Here we introduce the training algorithm of DGConv. Note that every DGConv layer is trained in the same way, implying that it can be easily plugged into a deep ConvNet by replacing the traditional convolution operations.

The training of DGConv can be simply implemented in existing software platforms such as PyTorch and TensorFlow. To see this, DGConv is computed by combining Eqn.\eqref{eq2}, \eqref{eq3}, \eqref{eq6}, and \eqref{eq4}.
All these equations define differentiable transformations except the sign function in Eqn.\eqref{eq4}. Therefore, the gradients from the loss function can be propagated down to the binary gates $g$ in Eqn.\eqref{eq6}, by simply using auto differentiation (AD) in the above platforms.
The only remaining thing to deal with is the sign function in Eqn.\eqref{eq4}.
The optimization of binary variables has been well established in the literature \cite{rastegari2016xnor, pmlr-v97-luo19a, luo2019switchable,shao2019ssn},  which can be also used to train DGConv. The gate params are optimized by Straight-Through Estimator similar to recent network quantization approaches, which is guaranteed to converge \cite{yin2019understanding}.
%
%
Furthermore, Appendix B also provides the explicit gradient computations of DGConv, facilitating implementation of DGConv in the platforms without auto differentiation.


\subsection{Groupable Residual Networks}


    \label{sec:3.3}
\begin{figure}
\begin{center}
   \includegraphics[width=1.0\linewidth]{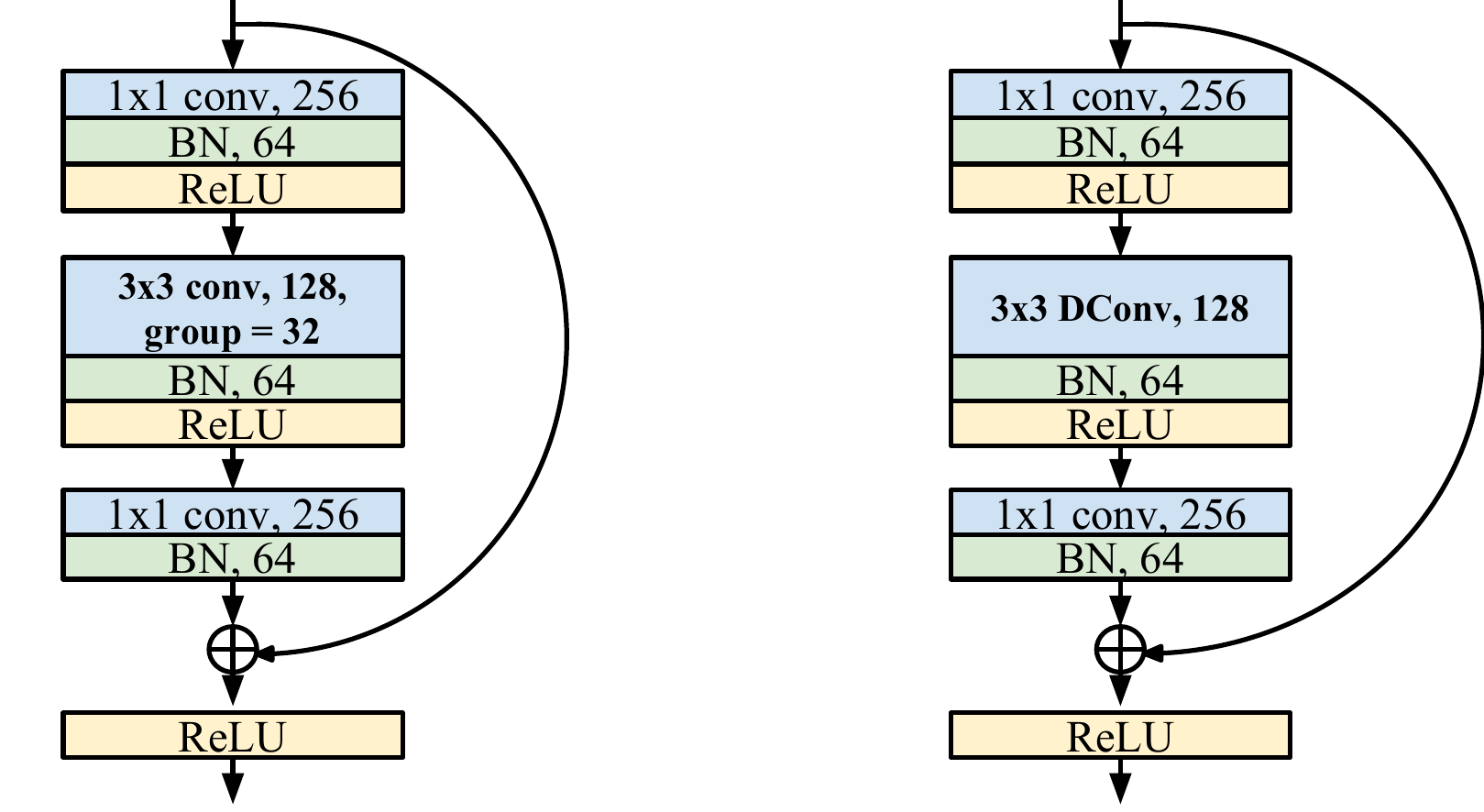}
    \caption{
    Comparison of the residual building blocks of\textit{ ResNeXt50$\_$32$\times$4d} (left) and \textit{Groupable-ResNeXt50} (right).  We simply replace all group convolution layers with dynamic grouping  convolution layers.
    }
    \label{fig:block}
\end{center}
\end{figure}

    DGConv is closely related to ResNet and ResNeXt, where ResNeXt extends ResNet by dividing channels into groups. DGConv can be also used with residual learning by simply replacing the traditional group convolutions of ResNeXt with the proposed dynamic grouping  convolutions, as shown in Fig.~\ref{fig:block}. We name this new network architecture Groupable ResNeXt.
    Table~\ref{tab:1} compares the architecture of Groupable-ResNeXt50 (G-ResNeXt50) to that of the original ResNeXt50.

\textbf{Resource-constrained Groupable Networks.}
Besides simply replacing convolution layers by using DGConv layers in a deep network, we also provide a resource-constrained training scheme. Different DGConv layers can have different group numbers, such that how and where to reduce computations are totally dependent on training data and tasks.

Towards this end, we propose a regularization term denoted by $\zeta$ to constrain the computational complexity of Groupable-ResNeXt, where $\zeta$ is computed by
    \begin{equation}
        \label{eq8}
        \begin{aligned}
        \zeta  = \sum_{\ell=1}^{L} ~\zeta_{\ell} ~~\mathrm{and}~~
        \zeta_\ell = \sum_{i=1}^{C^{\mathrm{in}}}\sum_{j=1}^{C^{\mathrm{out}}} u_{ij},~\forall u_{ij}\in U
        \end{aligned}
    \end{equation}
    where $L$ denotes the number of DGConv layers and $u_{ij}$ denotes an element of $U$. It is seen that $\zeta_\ell$ represents the number of non-zero elements in $U$, measuring the number of activated convolution weights (kernels) of the $\ell$-th DGConv layer. Thus, $\zeta$ can be treated as a measurement of the model's computational complexity.
In fact, it can be deduced by Eqn.\eqref{eq7} that the sum of each row or each column of $U$ can be calculated as $\prod_{k=1}^K(1+g_k)$. Substituting it to Eqn.\eqref{eq8} gives us
    \begin{equation}
        \label{eq9}
        \zeta = \sum_{\ell=1}^{L}\zeta_k = \sum_{\ell=1}^{L} C^\ell \cdot \prod_{k=1}^{K^\ell}(1+g^\ell_k),
    \end{equation}
    where $g^\ell_k$ and $K^\ell$ indicate $g_k$ and $K$ in the $\ell$-th layer, respectively. Here we assume $C^\ell=C^{\mathrm{in}}=C^{\mathrm{out}}$.
    Let $o$ represent the desire computational complexity of the entire network, our objective is to search a deep model that
    \begin{eqnarray*}
        &&\mathrm{minimize}~~\mathcal{L}(\{\omega_\ell\}_{\ell=1}^L,\{\tilde{g}_\ell\}_{\ell=1}^L)
        \cdot [\frac{o}{\zeta}]^{a},\\
        &&\text{subject to} ~~\zeta \leq o
    \end{eqnarray*}
    where $[\frac{o}{\zeta}]^{a}$ is a weighted product to approximate the Pareto optimal problem \cite{tan2018mnasnet} and $a$ is a constant value. We have $a=0$ if $\zeta \leq o$, implying that the complexity constraint is satisfied. Otherwise, $a=\alpha$ is used to penalize the model complexity when $\zeta>o$.
    For the value of $\alpha$, \cite{tan2018mnasnet} empirically set $\alpha=-1$ or $-0.07$ and this setting works well in reinforcement learning by using rewards. However, these empirical values make the regularizer too sensitive in our problem. In our experiments, we have $\alpha = -0.02$ as a constant.

    The above loss function can be optimized by using SGD. By setting the value of $o$,  we can learn deep neural networks under different complexity constraints, allowing us to carry on careful studies on the trade-off between model accuracy and computational complexity.

\begin{table}
        \scalebox{0.65}{
        \begin{tabular}{l|c|c|c}
        \hline
        stage & output & ResNeXt50$\_$32x4d & G-ResNeXt50 \\
        \hline\hline
        conv1 & $112\times112$ & $7\times7$, 64, stride 2 & $7\times7$, 64, stride 2 \\
        maxpool & $56\times56$ & $3\times 3$, stride 2 & $3\times3$, stride 2 \\
        \hline
        conv2 & $56\times56$ & $\begin{bmatrix}
        1\times1, 128\\
        3\times3, 128 ~~G=32\\
        1\times1, 256
        \end{bmatrix} \times 3$ & $\begin{bmatrix}
        1\times1, 128\\
        3\times3, 128 ~~DGConv\\
        1\times1, 256
        \end{bmatrix} \times 3$\\
        \hline
        conv3 & $28\times28$ & $\begin{bmatrix}
        1\times1, 256\\
        3\times3, 256 ~~G=32\\
        1\times1, 512
        \end{bmatrix} \times 4$  &  $\begin{bmatrix}
        1\times1, 256\\
        3\times3, 256 ~~DGConv\\
        1\times1, 512
        \end{bmatrix} \times 4$ \\
        \hline
        conv4 & $14\times14$ &  $\begin{bmatrix}
        1\times1, 512\\
        3\times3, 512 ~~G=32\\
        1\times1, 1024
        \end{bmatrix} \times 6$ &  $\begin{bmatrix}
        1\times1, 512\\
        3\times3, 512 ~~DGConv\\
        1\times1, 1024
        \end{bmatrix} \times 6$\\
        \hline
        conv5 & $7\times7$ & $\begin{bmatrix}
        1\times1, 1024\\
        3\times3, 1024 ~~G=32\\
        1\times1, 2048
        \end{bmatrix} \times 3$ & $\begin{bmatrix}
        1\times1, 1024\\
        3\times3, 1024 ~~DGConv\\
        1\times1, 2048
        \end{bmatrix} \times 3$\\
        \hline
        \end{tabular}}
        \vspace{3pt}
        \caption{Comparison of network structures  between \textit{ ResNeXt50$\_$32$\times$4d} and \textit{Groupable-ResNeXt50}. In \textit{ ResNeXt50$\_$32$\times$4d}, $G=32$ is a hyper-parameter, indicating group number in channel domain.  \textit{Groupable-ResNeXt50} replaces all group convolution layers in \textit{ResNeXt50$\_$32$\times$4d} by using DGConv layers, keeping others unchanged.}
        \label{tab:1}

\end{table}

\begin{figure*}[t]
\begin{center}
   \includegraphics[width=0.75\linewidth]{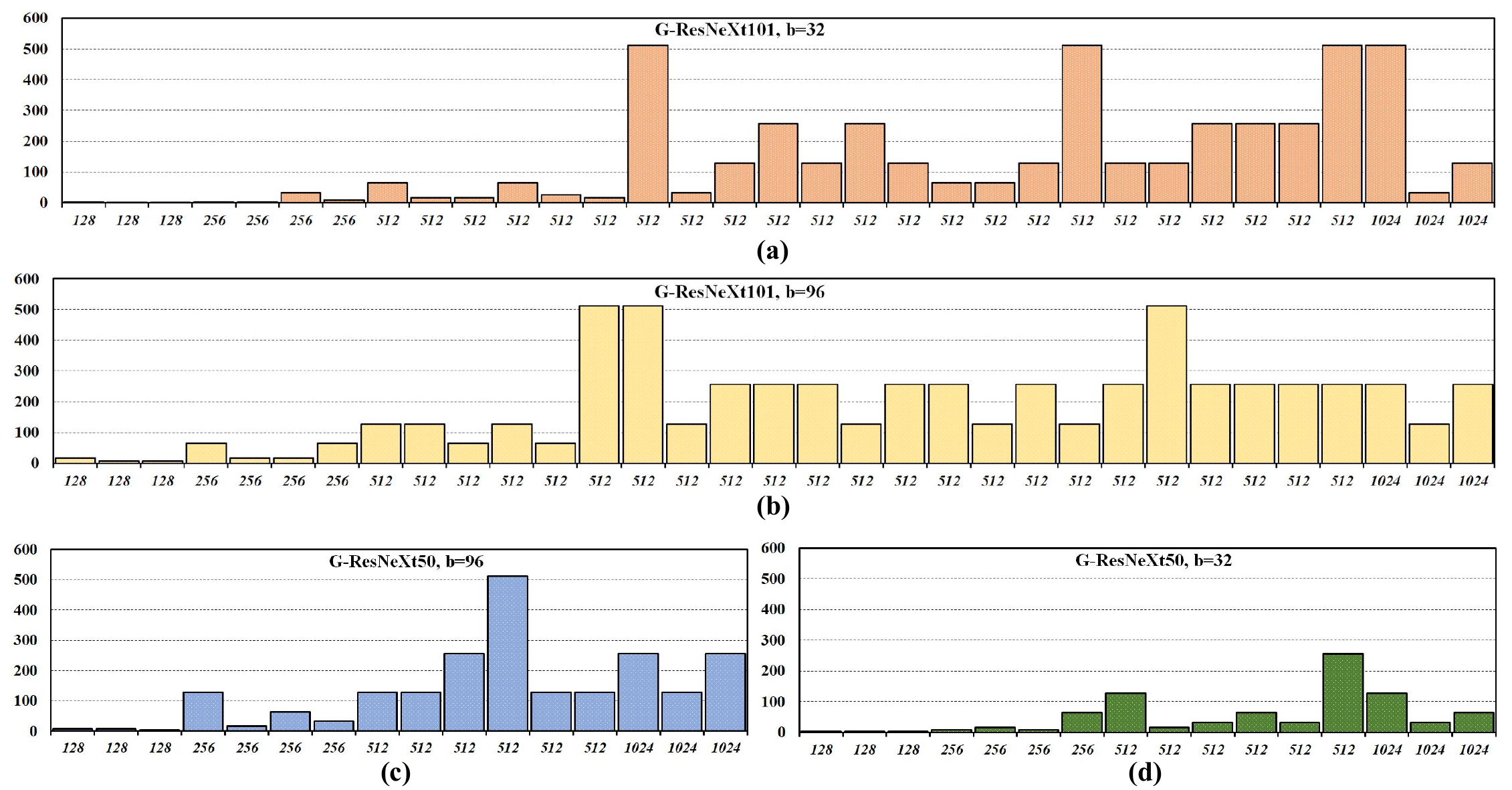}
    \caption{
     Learned number of groups for each DGConv layer in Groupable-ResNeXt, including: (a) G-ResNeXt101, $b=32$, (b) G-ResNeXt101, $b=96$, (c) G-ResNeXt50, $b=32$ and (d) G-ResNeXt50, $b=96$. The $x$-axis denotes the number of channels in DGConv layers under network's input to output direction, and the $y$-axis is the group number of corresponding layers.
     }
     \label{fig:4}
\end{center}
\end{figure*}

\section{Experiments}

\textbf{Implementation.}
\label{sec:4.1}
We conduct experiments on the challenging ImageNet~\cite{deng2009imagenet} benchmark, which has 1.2 million images for training and 50k images for validation.
Following Section~\ref{sec:3.3} and ~\cite{xie2017aggregated}, we construct 50-layer and 101-layer Groupable ResNeXts.
%
In the training stage, each input image is of size $224\times224$ that is randomly cropped from randomly horizontal flipped. The overall batch size is 512, partitioned to 16 GPUs (32 samples per GPU).  We train the networks by using SGD with momentum $0.9$ and weight decay $1e^{-4}$.
We adopt the cosine learning rate schedule~\cite{loshchilov2016sgdr} and weight initialization of ~\cite{he2015delving}.
In the evaluation stage, the error is evaluated on a single $224\times224$ center crop.
For Groupable ConvNets, the continuous gates $\widetilde{g}$ are the only extra parameters required to train. We initialize them as small values $1e^{-8}$ or $-1e^{-8}$ randomly.

\textbf{Resource Constraint.}
In experiments, we derive the resource constraint $o$ by $o = \frac{\sum_{\ell=1}^{L}C_{\ell}^{2}}{b}$,
where $\mathbf{b}$ denotes a scale of complexity of the group convolution layers in the entire network.
For an example, when $b=32$, $\frac{\sum_{\ell=1}^{L}C_{\ell}^{2}}{b}$ is equivalent to the number of parameters of all GConv layers in ResNeXt $32\times4$d, and $o$ represents the complexity of GConv layers in ResNeXt $32\times4$d.
When $b=64$, $o$ is $0.5\times$ complexity compared to the ResNeXt $32\times4$d, and so on.
By setting $b$, we are able to control the overall complexity of Groupable ConvNets.

\begin{table}
\centering
\small
\scalebox{0.7}{
        \begin{tabular}{l|c|c}
        \hline
        Architecture & Params\#  & Top-1 Accuracy  \\
        \hline
        ResNet50 & 25 M &  76.4  \\
        InceptionV3 & 23 M & 77.5 \\
        IBN-Net50-a & 25 M & 77.5 \\
        SE-ResNet50 & 28 M & 77.7 \\
        ResNeXt50  & \textbf{25 M} & \textbf{77.8} \\
        DenseNet161(k=48)  & 29 M  & 77.8 \\
        DenseNet264(k=32)&  33 M &  77.9  \\ 
        G-ResNeXt50(b=32, ours) & \textbf{25M} & \textbf{78.4}\\ \hline
        ResNet101 & 44 M &  78.0  \\
        SE-ResNet101  & 48 M & 78.4 \\
        ResNeXt101  & \textbf{44 M} & \textbf{78.8} \\
        DenseNet-232 (k=48) &  55 M  & 78.8 \\
        G-ResNeXt101(b=32, ours) & \textbf{43M} & \textbf{79.9}\\
        \hline
        \end{tabular}
        \vspace{3pt}

}
        \caption{Comparisons of top-1 and top-5 accuracy on ImageNet when the number of \#parameters in different networks are almost the same. Our approach shows superior performance to its counterparts. Groupable-ResNeXt is abbreviated as G-ResNeXt. The accuracy is evaluated on a signle $224\times224$ crop of image.
       We set scale constant b of the model complexity in Groupable-ResNeXt to $32$, so as to keep proximate parameter size with their counterpart ResNet and ResNeXt. We choose ResNeXt of setting $32\times4$d, which outperforms other settings in ~\cite{xie2017aggregated}  }
        \label{tab:2}
\end{table}


\textbf{Comparisons.}
We first evaluate the performance of Groupable-ResNeXt and its counterparts ResNet/ResNeXt. For fair comparison, we re-implement ResNet and ResNeXt under the settings of Section.~\ref{sec:4.1}, achieving comparable results to the original papers (\eg top-1 accuracy of ResNeXt101, $32\times4$d,  79.1\% (ours) \vs 78.8\%\cite{xie2017aggregated} ).
Table~\ref{tab:2} shows the results, and Fig.~\ref{fig:4} shows the learned group numbers.
Although maintaining similar module topology as ResNeXt, Groupable-ResNeXt learns optimal grouping  strategies for group convolution. 
Compared to ResNet50 and ResNeXt50, G-ResNeXt50 obtains 1.5\% / 0.5\% higher top-1 accuracy. This trend is also observed in deeper architectures ResNet101 and ResNeXt101, and the gains of top-1 accuracy are enlarged to 1.7\% and 0.8\%.

Fig.\ref{fig:2} and Fig.\ref{fig:4} show the learned group numbers. Table \ref{tab:2} reports performance of G-ResNext50($b=32$) and G-ResNeXt101($b=32$), which correspond to Fig.\ref{fig:4} (d) and Fig.\ref{fig:4} (a).
Unlike ResNeXt that shares uniform group number, diverse group numbers could be observed in G-ResNext.
An interesting phenomenon is that different networks manifest some homology.
That is, \textit{when preserving the overall model complexity, DGConv tends to allocate more computation in lower layers.}
This is an evidence that the representation ability of ConvNet is highly related to the design of lower layers.
%

He \etal~\cite{xie2017aggregated} found that, when the network complexity is similar, the networks with larger cardinality perform better than those deeper or wider. The performance gain comes from stronger representations. We suggest that the representations could be even stronger by adjusting the grouping  strategy at each layer using DGConv.

\begin{figure}
\begin{center}
  \includegraphics[width=1\linewidth]{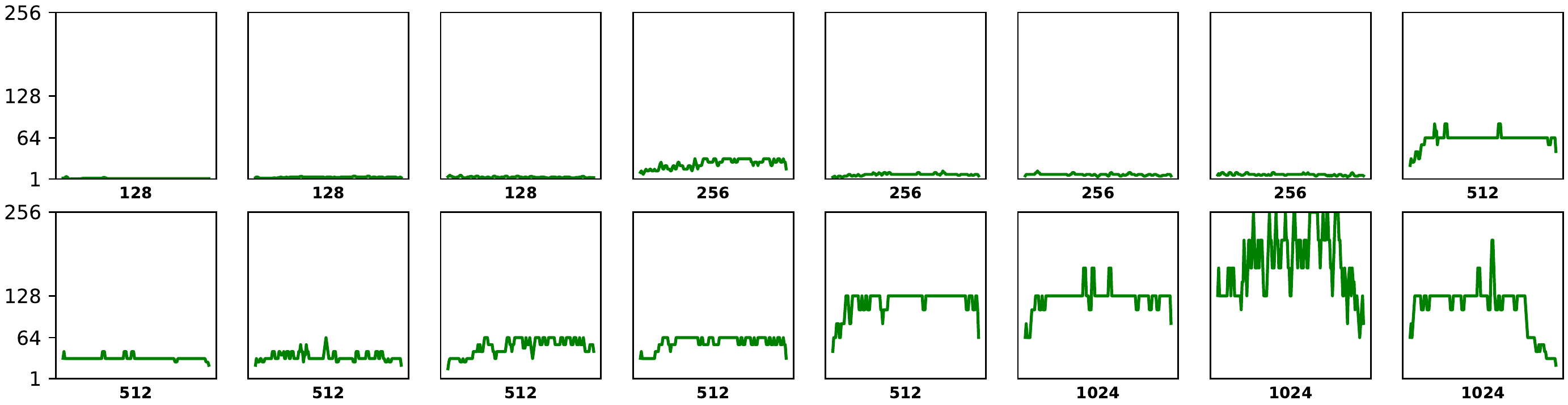}
  \small{(a) Learning dynamics of group Number in different layers}
  \includegraphics[width=1\linewidth]{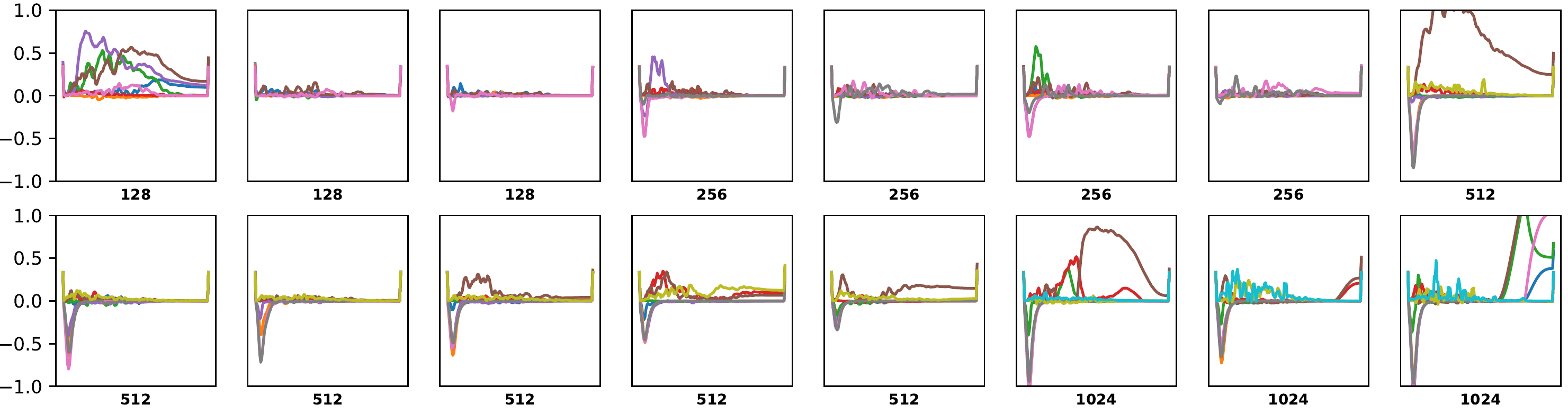}
  \small{(b) Learning dynamics of gate values $\tilde{g}$ in different layers}
 \end{center}
    \caption{Learning dynamics of group number and learnable gate vector $\tilde{g}$ during training Groupable-ResNeXt50 on ImageNet. (a) visualizes how the number of group in different depth evolves with training. (b) shows the corresponding learning process of gate values $\tilde{g}$. The number of channels is plotted for each layer (in the bottom).} 
    \label{fig:5}
\end{figure}

\textbf{Learning dynamics of DGConv.}
For every DGConv layers in G-ResNeXt50 ($b=32$), we plot the learning procedure of group numbers and value of gates $g$ in Fig.~ \ref{fig:5}. To our observation, DGConv appears some features. First, different DGConv layer shows different learning dynamics. Second, similar to Fig.~\ref{fig:4}, lower layers prefer fewer groups than higher layers. Therefore, lower layers tend to have fewer groups corresponding to more parameters, implying that they are essential for extracting texture-related features.

\textbf{Complexity \vs Accuracy.}
The resource constraint $o$ allows us to learn optimal grouping  strategies subject to a given model complexity threshold. 
We then explore the trade-off between complexity of group convolution and model accuracy. 
Table \ref{tab:3} shows our results, where ``FLOPs'' denotes computational complexity of all group convolution layers in a network. We set the FLOPs of ResNeXt as baseline and show complexity of Groupable-ResNeXt by proportion. By modifying $b$, we alter the constraint $o$ and learn  Groupable-ResNeXt of various capacity.
For example, when $b=64$, $o$ is equivalent to the size of group convolutions with group number $64$ uniformly, and Groupable-ResNeXt will be regularized to choose group strategy less than $0.5\times$ ResNeXt's complexity. 

From Table \ref{tab:3}, we see that G-ResNeXt50 achieves comparable top-1 accuracy with ResNeXt50 in the $b=96$ setting, and G-ResNeXt101 achieves comparable top-1 accuracy with ResNeXt101 in the $b=256$ setting. These results indicate that DGConv is able to learn more efficient group strategy than regular GConv when preserving accuracy. He \etal~\cite{xie2017aggregated} suggests that learning wide cardinality has stronger representation than wide depth or width, and we learn dynamic grouping  to improve representation learning of wide cardinality.

Furthermore, we also see the strong robustness of dynamic grouping  convolution, even when the computational complexity of group convolution is significantly reduced. For example, when FLOPs decrease from $0.70\times$ to $0.47\times$, G-ResNeXt101 is able to preserve its accuracy (about $79.8$\% top-1 accuracy).
\begin{table}
        \centering
        \small
        \scalebox{0.8}{
        \begin{tabular}{l|l|c| c|c}
        \hline
        Architecture & Settings & GConv FLOPs & top-1 & top-5  \\
        \hline
        ResNeXt50 & $32\times4$d & $1.00\times$ & 77.9 & 93.9  \\
        G-ResNeXt50 & $b=32$ & $0.83\times$ & 78.4 & 94.0  \\
        G-ResNeXt50 &  $b=48$ & $0.61\times$ & 78.2 & 93.9  \\
        G-ResNeXt50 &  $b=64$ & $0.39\times$ & 78.0 & 93.9  \\
        G-ResNeXt50 &  $b=96$ & $0.27\times$ & 78.0 & 93.9  \\
        G-ResNeXt50 &  $b=128$ &  $0.20$ & 77.8 &  93.8  \\
        \hline
        ResNeXt101 & $32\times4$d  & $1.00\times$ & 79.1 & 94.2  \\
        G-ResNeXt101 &  $b=32$ & $0.70\times$ & 79.9 & 94.7  \\
        G-ResNeXt101 &  $b=48$ & $0.58\times$  & 79.7 & 94.6  \\
        G-ResNeXt101 &  $b=64$ & $0.47\times$ & 79.8 & 94.7  \\
        G-ResNeXt101 &  $b=96$ & $0.22\times$ & 79.5 & 94.5  \\
        G-ResNeXt101 &  $b=128$ & $0.22\times$ & 79.4 & 94.5  \\
        G-ResNeXt101 &  $b=256$ & $0.14\times$ & 79.0 & 94.3  \\
        \hline
        \end{tabular}}
        \vspace{3pt}
        \caption{Trade-off between complexity and accuracy. Here GConv FLOPs represents the computational complexity of all group convolution layers in the corresponding network architecture. The FLOPs of ResNeXt50/101 is regarded as baselines, and we report complexity of other models as proportions of them. All G-ResNeXt models outperform baselines at top1 accuracy with much less computation. Even given only about $\sim1/4$ FLOPs, both G-ResNeXt50/101 achieve comparable top1/top5 accuracy with respect to baselines.}
        \label{tab:3}
\end{table}


\textbf{Deeper or Wider Networks.}
Next we extend our experiments to more complex networks. We expand ResNet101 to $\sim2\times$ complexity by increasing its width, depth, and cardinality respectively. When expanding on cardinality, we implement both the regular GConv and DGConv. Table \ref{tab:4} reports our results. The larger ResNet and ResNeXt are implemented by following~\cite{xie2017aggregated, he2016identity}. G-ResNeXt101 is constrained to the size of ResNeXt101 $2\times64$d. In Table \ref{tab:4}, we see that increasing the model complexity consistently improves network performance (\eg the original ResNet101 is $78.2$\%). Besides, increasing cardinality brings larger improvement than increasing the network depth and width (\eg $79.8$\%/$79.6$\%/$80.1$\% \vs $78.6$\%/$78.8$\%). Among the last three networks with larger cardinality, G-ResNeXt101 ($b=2$) outperforms corresponding ResNext101 ($2\times64d$) by $0.5$\% top-1 accuracy. G-ResNeXt101 increases cardinality by using DGConv. We show that DGConv is superior to regular GConv even in more complex networks.

\begin{table}
        \centering
        \footnotesize
        \begin{tabular}{l|c|c|c|c}
        \hline
        Architecture & Settings  & Complexity & top-1 & top-5  \\
        \hline
        ResNet200 (depth)& $1\times64$d & $2\times$ ResNet101  & 78.6  &  94.1   \\
        ResNet101 (wider~\cite{he2016identity}) & $1\times100$d &  $ 2\times$ ResNet101 &78.8 &  94.4 \\
        ResNeXt101 (card.) & $64\times4$d  &  $2\times$ ResNet101  &79.8 & 94.7 \\
        ResNeXt101 (card.) & $2\times64$d  & $2\times$ ResNet101 &79.6 & 94.6  \\
        G-ResNeXt101 (card.) & $b=2$ & $2\times$ ResNet101  & \textbf{80.1} &  \textbf{94.7} \\
        \hline
        \end{tabular}
        \vspace{3pt}
        \caption{Network performance on ImageNet when increasing number of parameters to $2\times$ ResNeXt101. All of above networks are re-implemented under the same settings for fair comparison. G-ResNeXt represents Groupable-ResNeXt. To keep proximate parameter size with ResNeXt101 $2\times64$d, the scale constant b of the model complexity in G-ResNeXt is set to $2$. G-ResNeXt achieves the highest top1/top5 accuracy among all architectures.}
        \label{tab:4}
\end{table}

\textbf{Reproducibility.}
We verify the reproducibility of DGConv. We retrain G-ResNeXt101 by maintaining training strategy and hyper-parameters, but initialize gates $g$ as $1\times10^{-8}$ or $-1\times10^{-8}$ randomly with different random seeds. 
We name the retrained models "G-ResNeXt101R2" and "G-ResNeXt101R3". Table \ref{tab:5} reports their performances. All models are trained with constraint $b=32$, showing comparable top-1 accuracy. These results indicate that DGConv is able to consistently express strong representation ability.
We also see that the learned models have similar performance with slightly different grouping  strategy, showing the flexibility of DGConv. Detailed group number distribution can be seen in  Appendix D.

\begin{table}
        \centering
        \footnotesize
        \begin{tabular}{l|l|c|c|c}
        \hline
        Architecture & Settings &$\#$Params  & top-1 & top-5  \\
        \hline
        G-ResNeXt101 &  $b=32$ & $43.3\times10^6$ & 79.9 &  94.7   \\
        G-ResNeXt101R2 &  $b=32$ &$43.8\times10^6$& 79.8 &  94.5 \\
        G-ResNeXt101R3 &  $b=32$ & $43.0\times10^6$ & 79.6 & 94.5 \\
        \hline
        \end{tabular}
        \vspace{3pt}
        \caption{Reproducibility experiments results. G-ResNeXt101R2 and G-ResNeXt101R3 are re-trained under the same setting as G-ResNeXt. After training, these three models approach proximate results and top1/top5 accuracy even they use different random seeds for initialization, which shows that DGConv is robust to randomness.}
        \label{tab:5}
\end{table}

\textbf{Evaluation of Learned Architecture}
We extend our experiments to the architecture learned by DGConv. We replace group numbers of each GConv layers in ResNeXt with the group numbers learned by G-ResNeXt. Then the formed models are directly trained on ImageNet \textbf{from scratch}. 
Table.~\ref{tab:6} reports their performance. As we can see, the ResNeXt models learned by DGConv perform comparable top-1 and top-5 accuracy with G-ResNeXt, superior to the $32\times4$d baseline. The results manifest strong representation in the learned structure.

\begin{table}
        \begin{center}
        \footnotesize
        \begin{tabular}{l|l|c|c}
        \hline
        Architecture & Settings  & top-1 & top-5  \\
        \hline
        ResNeXt50 &  $32\times4$d  & 77.9 & 93.9\\
        G-ResNeXt50 &  $b=32$  & 78.4 &  94.0   \\
        ResNeXt50$^*$ &  learned by $b=32$ &  \textbf{78.3 }& \textbf{94.0}   \\
        G-ResNeXt50 &  $b=96$ &   78.0 &  93.9  \\
        ResNeXt50$^*$ &  learned by $b=96$  &\textbf{ 78.0} & \textbf{93.9}  \\
        \hline
        ResNeXt101 &  $32\times4$d  & 79.1 & 94.2\\
        G-ResNeXt101 &  $b=32$  & 79.9 &  94.7   \\
        ResNeXt101$^*$ &  learned by $b=32$ &  \textbf{79.8}& \textbf{94.7}   \\
        G-ResNeXt101 &  $b=96$ &   79.5&  94.5  \\
        ResNeXt101$^*$ &  learned by $b=96$  &\textbf{ 79.5} & \textbf{94.5}  \\
        \hline
        \end{tabular}
        \end{center}
        \vspace{3pt}
        \caption{Performance of ResNeXt using group number learned by DGConv, denoted by ResNeXt$^*$. To demonstrate the effectiveness of the structures learned by DGConv, we just simply replace the group numbers in ResNeXt50 by the numbers learned from G-ResNeXt. 
        }
        \label{tab:6}
\end{table}

\section{Conclusion}
In this work, we propose a novel architecture Groupable ConvNet (GroupNet) for computation efficiency and performance boosting. GroupNet is able to differentiably learn group strategy for convolution operation on a layer-by-layer basis. It has been demonstrated that GroupNet outperforms ResNet and ResNeXt in terms of both accuracy and computational complexity. To achieve GroupNet, we develop dynamic grouping  convolution (DGConv), providing an unified representation for convolution operation. DGConv can be easily plugged into any deep network model and is expected to learn a better feature representation for convolution layer.

{
\section{Acknowledgement} 
 This work is supported in part by SenseTime Group Limited, and in part by the General Research Fund through the Research Grants Council of Hong Kong under Grants CUHK14202217, CUHK14203118, CUHK14205615, CUHK14207814, CUHK14213616. }

{\small
\bibliographystyle{ieee_fullname}
\bibliography{egbib}

\begin{thebibliography}{10}\itemsep=-1pt

\bibitem{baker2016designing}
Bowen Baker, Otkrist Gupta, Nikhil Naik, and Ramesh Raskar.
\newblock Designing neural network architectures using reinforcement learning.
\newblock {\em arXiv preprint arXiv:1611.02167}, 2016.

\bibitem{batselier2017constructive}
Kim Batselier and Ngai Wong.
\newblock A constructive arbitrary-degree kronecker product decomposition of
  tensors.
\newblock {\em Numerical Linear Algebra with Applications}, 24(5):e2097, 2017.

\bibitem{cai2017reinforcement}
Han Cai, Tianyao Chen, Weinan Zhang, Yong Yu, and Jun Wang.
\newblock Reinforcement learning for architecture search by network
  transformation.
\newblock {\em arXiv preprint arXiv:1707.04873}, 2017.

\bibitem{deng2009imagenet}
Jia Deng, Wei Dong, Richard Socher, Li-Jia Li, Kai Li, and Li Fei-Fei.
\newblock Imagenet: A large-scale hierarchical image database.
\newblock In {\em 2009 IEEE conference on computer vision and pattern
  recognition}, pages 248--255. Ieee, 2009.

\bibitem{yin2019understanding}
Yin~Penghang et al.
\newblock Understanding straight-through estimator in training activation
  quantized neural nets.
\newblock 2019.

\bibitem{he2015delving}
Kaiming He, Xiangyu Zhang, Shaoqing Ren, and Jian Sun.
\newblock Delving deep into rectifiers: Surpassing human-level performance on
  imagenet classification.
\newblock In {\em Proceedings of the IEEE international conference on computer
  vision}, pages 1026--1034, 2015.

\bibitem{he2016deep}
Kaiming He, Xiangyu Zhang, Shaoqing Ren, and Jian Sun.
\newblock Deep residual learning for image recognition.
\newblock In {\em Proceedings of the IEEE conference on computer vision and
  pattern recognition}, pages 770--778, 2016.

\bibitem{he2016identity}
Kaiming He, Xiangyu Zhang, Shaoqing Ren, and Jian Sun.
\newblock Identity mappings in deep residual networks.
\newblock In {\em European conference on computer vision}, pages 630--645.
  Springer, 2016.

\bibitem{howard2017mobilenets}
Andrew~G Howard, Menglong Zhu, Bo Chen, Dmitry Kalenichenko, Weijun Wang,
  Tobias Weyand, Marco Andreetto, and Hartwig Adam.
\newblock Mobilenets: Efficient convolutional neural networks for mobile vision
  applications.
\newblock {\em arXiv preprint arXiv:1704.04861}, 2017.

\bibitem{huang2018condensenet}
Gao Huang, Shichen Liu, Laurens Van~der Maaten, and Kilian~Q Weinberger.
\newblock Condensenet: An efficient densenet using learned group convolutions.
\newblock In {\em Proceedings of the IEEE Conference on Computer Vision and
  Pattern Recognition}, pages 2752--2761, 2018.

\bibitem{ioffe2015batch}
Sergey Ioffe and Christian Szegedy.
\newblock Batch normalization: Accelerating deep network training by reducing
  internal covariate shift.
\newblock {\em arXiv preprint arXiv:1502.03167}, 2015.

\bibitem{krizhevsky2012imagenet}
Alex Krizhevsky, Ilya Sutskever, and Geoffrey~E Hinton.
\newblock Imagenet classification with deep convolutional neural networks.
\newblock In {\em Advances in neural information processing systems}, pages
  1097--1105, 2012.

\bibitem{liu2018progressive}
Chenxi Liu, Barret Zoph, Maxim Neumann, Jonathon Shlens, Wei Hua, Li-Jia Li, Li
  Fei-Fei, Alan Yuille, Jonathan Huang, and Kevin Murphy.
\newblock Progressive neural architecture search.
\newblock In {\em Proceedings of the European Conference on Computer Vision
  (ECCV)}, pages 19--34, 2018.

\bibitem{liu2017hierarchical}
Hanxiao Liu, Karen Simonyan, Oriol Vinyals, Chrisantha Fernando, and Koray
  Kavukcuoglu.
\newblock Hierarchical representations for efficient architecture search.
\newblock {\em arXiv preprint arXiv:1711.00436}, 2017.

\bibitem{liu2018darts}
Hanxiao Liu, Karen Simonyan, and Yiming Yang.
\newblock Darts: Differentiable architecture search.
\newblock {\em arXiv preprint arXiv:1806.09055}, 2018.

\bibitem{loshchilov2016sgdr}
Ilya Loshchilov and Frank Hutter.
\newblock Sgdr: Stochastic gradient descent with warm restarts.
\newblock {\em arXiv preprint arXiv:1608.03983}, 2016.

\bibitem{luo2019switchable}
Ping Luo, Ruimao Zhang, Jiamin Ren, Zhanglin Peng, and Jingyu Li.
\newblock Switchable normalization for learning-to-normalize deep
  representation.
\newblock {\em IEEE Transactions on Pattern Analysis and Machine Intelligence},
  2019.

\bibitem{pmlr-v97-luo19a}
Ping Luo, Peng Zhanglin, Shao Wenqi, Zhang Ruimao, Ren Jiamin, and Wu Lingyun.
\newblock Differentiable dynamic normalization for learning deep
  representation.
\newblock In Kamalika Chaudhuri and Ruslan Salakhutdinov, editors, {\em
  Proceedings of the 36th International Conference on Machine Learning},
  volume~97 of {\em Proceedings of Machine Learning Research}, pages
  4203--4211, Long Beach, California, USA, 09--15 Jun 2019. PMLR.

\bibitem{ma2018shufflenet}
Ningning Ma, Xiangyu Zhang, Hai-Tao Zheng, and Jian Sun.
\newblock Shufflenet v2: Practical guidelines for efficient cnn architecture
  design.
\newblock In {\em Proceedings of the European Conference on Computer Vision
  (ECCV)}, pages 116--131, 2018.

\bibitem{miikkulainen2019evolving}
Risto Miikkulainen, Jason Liang, Elliot Meyerson, Aditya Rawal, Daniel Fink,
  Olivier Francon, Bala Raju, Hormoz Shahrzad, Arshak Navruzyan, Nigel Duffy,
  et~al.
\newblock Evolving deep neural networks.
\newblock In {\em Artificial Intelligence in the Age of Neural Networks and
  Brain Computing}, pages 293--312. Elsevier, 2019.

\bibitem{pham2018efficient}
Hieu Pham, Melody~Y Guan, Barret Zoph, Quoc~V Le, and Jeff Dean.
\newblock Efficient neural architecture search via parameter sharing.
\newblock {\em arXiv preprint arXiv:1802.03268}, 2018.

\bibitem{rastegari2016xnor}
Mohammad Rastegari, Vicente Ordonez, Joseph Redmon, and Ali Farhadi.
\newblock Xnor-net: Imagenet classification using binary convolutional neural
  networks.
\newblock In {\em European Conference on Computer Vision}, pages 525--542.
  Springer, 2016.

\bibitem{real2018regularized}
Esteban Real, Alok Aggarwal, Yanping Huang, and Quoc~V Le.
\newblock Regularized evolution for image classifier architecture search.
\newblock {\em arXiv preprint arXiv:1802.01548}, 2018.

\bibitem{real2017large}
Esteban Real, Sherry Moore, Andrew Selle, Saurabh Saxena, Yutaka~Leon Suematsu,
  Jie Tan, Quoc~V Le, and Alexey Kurakin.
\newblock Large-scale evolution of image classifiers.
\newblock In {\em Proceedings of the 34th International Conference on Machine
  Learning-Volume 70}, pages 2902--2911. JMLR. org, 2017.

\bibitem{sandler2018mobilenetv2}
Mark Sandler, Andrew Howard, Menglong Zhu, Andrey Zhmoginov, and Liang-Chieh
  Chen.
\newblock Mobilenetv2: Inverted residuals and linear bottlenecks.
\newblock In {\em Proceedings of the IEEE Conference on Computer Vision and
  Pattern Recognition}, pages 4510--4520, 2018.

\bibitem{shao2019ssn}
Wenqi Shao, Tianjian Meng, Jingyu Li, Ruimao Zhang, Yudian Li, Xiaogang Wang,
  and Ping Luo.
\newblock Ssn: Learning sparse switchable normalization via sparsestmax.
\newblock In {\em Proceedings of the IEEE Conference on Computer Vision and
  Pattern Recognition}, pages 443--451, 2019.

\bibitem{szegedy2017inception}
Christian Szegedy, Sergey Ioffe, Vincent Vanhoucke, and Alexander~A Alemi.
\newblock Inception-v4, inception-resnet and the impact of residual connections
  on learning.
\newblock In {\em Thirty-First AAAI Conference on Artificial Intelligence},
  2017.

\bibitem{szegedy2015going}
Christian Szegedy, Wei Liu, Yangqing Jia, Pierre Sermanet, Scott Reed, Dragomir
  Anguelov, Dumitru Erhan, Vincent Vanhoucke, and Andrew Rabinovich.
\newblock Going deeper with convolutions.
\newblock In {\em Proceedings of the IEEE conference on computer vision and
  pattern recognition}, pages 1--9, 2015.

\bibitem{szegedy2016rethinking}
Christian Szegedy, Vincent Vanhoucke, Sergey Ioffe, Jon Shlens, and Zbigniew
  Wojna.
\newblock Rethinking the inception architecture for computer vision.
\newblock In {\em Proceedings of the IEEE conference on computer vision and
  pattern recognition}, pages 2818--2826, 2016.

\bibitem{tan2018mnasnet}
Mingxing Tan, Bo Chen, Ruoming Pang, Vijay Vasudevan, and Quoc~V Le.
\newblock Mnasnet: Platform-aware neural architecture search for mobile.
\newblock {\em arXiv preprint arXiv:1807.11626}, 2018.

\bibitem{Wang_2019_CVPR}
Xijun Wang, Meina Kan, Shiguang Shan, and Xilin Chen.
\newblock Fully learnable group convolution for acceleration of deep neural
  networks.
\newblock In {\em The IEEE Conference on Computer Vision and Pattern
  Recognition (CVPR)}, June 2019.

\bibitem{xie2017genetic}
Lingxi Xie and Alan Yuille.
\newblock Genetic cnn.
\newblock In {\em Proceedings of the IEEE International Conference on Computer
  Vision}, pages 1379--1388, 2017.

\bibitem{xie2017aggregated}
Saining Xie, Ross Girshick, Piotr Doll{\'a}r, Zhuowen Tu, and Kaiming He.
\newblock Aggregated residual transformations for deep neural networks.
\newblock In {\em Proceedings of the IEEE Conference on Computer Vision and
  Pattern Recognition}, pages 1492--1500, 2017.

\bibitem{xie2018snas}
Sirui Xie, Hehui Zheng, Chunxiao Liu, and Liang Lin.
\newblock Snas: stochastic neural architecture search.
\newblock {\em arXiv preprint arXiv:1812.09926}, 2018.

\bibitem{zhang2017interleaved}
Ting Zhang, Guo-Jun Qi, Bin Xiao, and Jingdong Wang.
\newblock Interleaved group convolutions.
\newblock In {\em Proceedings of the IEEE International Conference on Computer
  Vision}, pages 4373--4382, 2017.

\bibitem{zhang2018shufflenet}
Xiangyu Zhang, Xinyu Zhou, Mengxiao Lin, and Jian Sun.
\newblock Shufflenet: An extremely efficient convolutional neural network for
  mobile devices.
\newblock In {\em Proceedings of the IEEE Conference on Computer Vision and
  Pattern Recognition}, pages 6848--6856, 2018.

\bibitem{zoph2016neural}
Barret Zoph and Quoc~V Le.
\newblock Neural architecture search with reinforcement learning.
\newblock {\em arXiv preprint arXiv:1611.01578}, 2016.

\bibitem{zoph2018learning}
Barret Zoph, Vijay Vasudevan, Jonathon Shlens, and Quoc~V Le.
\newblock Learning transferable architectures for scalable image recognition.
\newblock In {\em Proceedings of the IEEE conference on computer vision and
  pattern recognition}, pages 8697--8710, 2018.

\end{thebibliography}
}

\end{document}